\documentclass[letterpaper, 10 pt, conference]{ieeeconf}  

\IEEEoverridecommandlockouts                              

\usepackage{graphics}
\usepackage{epsfig} 
\usepackage{mathptmx} 
\usepackage{times} 
\usepackage{amsmath} 
\usepackage{amssymb}  
\usepackage{multirow}
\usepackage{url}
\usepackage{caption}
\usepackage{subcaption}
\usepackage[table]{xcolor}
\usepackage{import}
\usepackage{hyperref}
\usepackage{booktabs}
\usepackage{tabularx}
\usepackage{microtype}
\usepackage{gensymb} 
\usepackage{cite}
\usepackage{stfloats}
\usepackage{bm}
\usepackage{siunitx}
\usepackage{acro}
\DeclareAcronym{com}{ 
    short = {CoM}, 
    long  = {Center of Mass},
    first-style = short-long,
}
\DeclareAcronym{cc}{ 
    short = {CC}, 
    long  = {Center of Geometry},
    first-style = short-long,
}
\DeclareAcronym{cop}{ 
    short = {CoP}, 
    long  = {Center of Pressure},
    first-style = short-long,
}
\DeclareAcronym{wrt}{ 
    short = {w.r.t.}, 
    long  = {with respect to},
    first-style = short-long,
}
\DeclareAcronym{ee}{ 
    short = {EE}, 
    long  = {end-effector},
    first-style = short,
}
\DeclareAcronym{dof}{ 
    short = {DoF}, 
    long  = {Degree of Freedom},
    first-style = short-long,
}

\title{\LARGE \bf  Passive Aligning Physical Interaction of Fully-Actuated Aerial Vehicles for Pushing Tasks
}

\author{Tong Hui$^1$, Eugenio Cuniato$^2$, Michael Pantic$^2$, Marco Tognon$^3$, Matteo Fumagalli$^1$, Roland Siegwart$^2$
\thanks{This work has been supported by the European Unions Horizon 2020 Research and Innovation Programme AERO-TRAIN under Grant Agreement No. 953454. Corresponding author email: {\tt\small tonhu@dtu.dk}}
\thanks{$^1$ Department of Electrical and Photonics Engineering, Technical University of Denmark, Denmark}
\thanks{$^2$ Autonomous Systems Lab, ETH Zurich, Switzerland}
\thanks{$^3$ Univ Rennes, CNRS, Inria, IRISA, F-35000 Rennes, France}
}

\begin{document}

\maketitle
\thispagestyle{empty}
\pagestyle{empty}

\begin{abstract}
Recently, the utilization of aerial manipulators for
performing pushing tasks in non-destructive testing (NDT)
applications has seen significant growth. Such operations entail physical interactions between the aerial robotic system and the environment. End-effectors with multiple contact points are often used for placing NDT sensors in contact with a surface to be inspected. Aligning the NDT sensor and the work surface while preserving contact, requires that all available contact points at the end-effector tip are in contact with the work surface. 
With a standard full-pose controller, attitude errors often occur due to perturbations caused by modeling uncertainties, sensor noise, and environmental uncertainties. Even small attitude errors can cause a loss of contact points between the end-effector tip and the work surface. To preserve full alignment amidst these uncertainties, we propose a control strategy which selectively deactivates angular motion control and enables direct force control in specific directions. In particular, we derive two essential conditions to be met, such that the robot can passively align with flat work surfaces achieving full alignment through the rotation along non-actively controlled axes. Additionally, these conditions serve as hardware design and control guidelines for effectively integrating the proposed control method for practical usage. Real world experiments are conducted to validate both the control design and the guidelines. 
\end{abstract}

\section{Introduction} \label{sec:introduction}
In recent years, the utilization of aerial manipulators for performing pushing tasks in non-destructive testing (NDT) applications has seen significant growth ~\cite{Anibal2021,watson2022}. Substantial research efforts have been directed towards developing aerial manipulation platforms for such applications.
Among various types of platforms, fully-actuated aerial vehicles capable of generating 6-\ac{dof} forces and torques (i.e., wrenches) through thrust vectoring have gained prominence~\cite{hamandi2021, rashad2020}. 

In general, specialized \ac{ee}s (End-Effector) are designed for specific interaction tasks. In NDT applications, \ac{ee}s with multiple contact points are often used for placing contact-based sensors to collect measurements from the work surface~\cite{watson2022, trujillo2019}. To obtain accurate measurements from the NDT sensor, it is essential to ensure full alignment between the NDT sensor and the work surface~\cite{ndt} while being in contact. This can be achieved by having full contact between the \ac{ee} tip and the work surface, where all available contact points on the \ac{ee} tip are in contact with the work surface. On the other hand, undesired contact arises when there is a lack of contact between at least one available contact point and the work surface.

Various interaction control methods for aerial manipulation have previously been implemented~\cite{Anibal2021}. Among these methods, impedance control~\cite{bodie2019, rashad2019, fabio2014, sch2022, zhang2022, benzi2022} and hybrid motion/force control~\cite{hui2023, hao2023, Bodie2020, nava2020, cuniato} stand out as prevalent approaches in aerial manipulation tasks. When using the aforementioned control methods with fully-actuated aerial vehicles~\cite{bodie2019,Bodie2020}, the full alignment between the \ac{ee} and the work surface depends on the accuracy of the attitude control and state estimation. However, this can be challenging due to common uncertainties in aerial systems caused by modeling inaccuracies, mechanical imperfections, sensor noise, aerodynamic ceiling effects \cite{ceiling}, or unknown surface orientation. The resulted attitude errors from uncertainties cause undesired contact between the \ac{ee} tip and the work surface. Consequently, the loss of contact points leads to reduced support area~\cite{ghee1968} \ac{wrt} the full-contact scenario. This results in two possible scenarios: 1) \textit{Stable Near-Zero Dynamics} - It may remain stable during the interaction but fails to fully align with the work surface; 2) \textit{Instability} - It may capsize around the reduced support area. 
\begin{figure}[t]
   \centering
\includegraphics[width=\columnwidth]{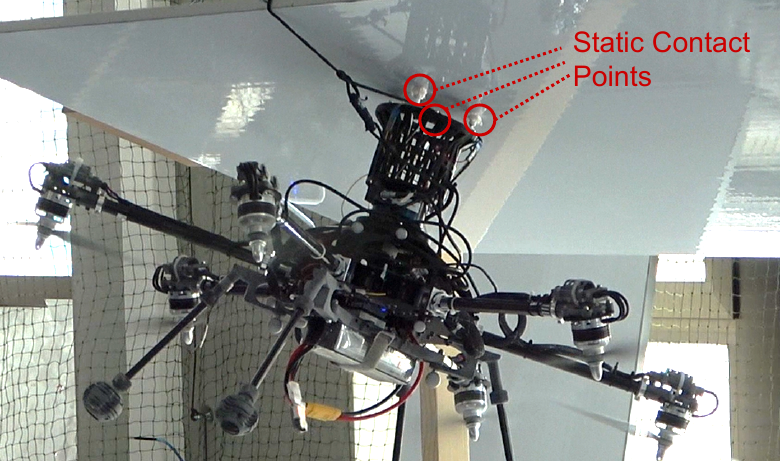}
    \caption{A fully-actuated aerial vehicle is pushing on an oriented flat work surface with static contact points at the \ac{ee} tip.}
    \label{fig:action}
\end{figure}

This work addresses the practical challenges of maintaining stable full alignment between the \ac{ee} of a fully-actuated aerial vehicle and the work surface during pushing, with the goal of guaranteeing reliable and robust aerial physical interaction for NDT applications. Taking inspiration from bipedal passive walking robots \cite{mc1990}, we propose a hybrid motion/force controller which selectively deactivates angular motion control and enables direct force control in specific directions. In particular, we derive two essential conditions, that if met, guarantee a passive alignement with flat work surfaces by leveraging passive dynamics along non-actively controlled axes.
These conditions serve as valuable guidelines for hardware and control design, facilitating the successful integration of the proposed control method into practical applications. In the end, we validate our findings with a fully-actuated aerial vehicle stably pushing on differently oriented work surfaces with static contact points, as in Fig.~\ref{fig:action}.
\section{Passive Aligning Physical Interaction}\label{sec:contact}
\begin{figure}[t]
   \centering
   \begin{subfigure}{0.48\columnwidth}
\includegraphics[width=0.9\columnwidth]{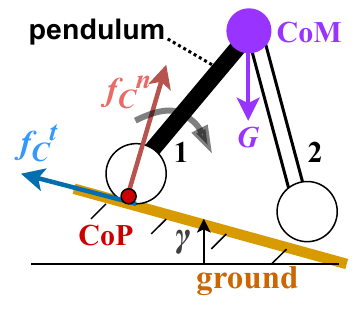}
    \caption{Passive walking.}
    \label{fig:passive_walking}
    \end{subfigure} \hfill
    \begin{subfigure}{0.48\columnwidth}
    \includegraphics[width=\columnwidth]{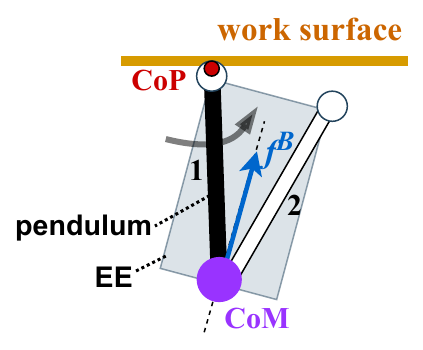}
    \caption{Passive aligning.}
    \label{fig:planr}
\end{subfigure}
\caption{Planar schematics of the pendulum model for passive walking (a) and passive aligning physical interaction (b).}\label{fig:inspiration}
\end{figure}

In this section, we introduce the fundamental assumptions for obtaining the passive aligning behaviour during physical interactions. In the studies on passive walking robots, the planar system model with two legs can be simplified as two rigid links being connected to the system \ac{com} \cite{mc1990}, see Fig.~\ref{fig:passive_walking}. When one link is in contact with the ground at the \ac{cop}, the gravity force vector $\bm{G}$ causes a torque around the \ac{cop} when it does not pass through the \ac{cop} with the non zero slope $\gamma$. When there is no slipping at the current contact point (i.e., \ac{cop}), the system rotates around the \ac{cop} like a simple pendulum until the other link gets in contact with the ground. This behaviour entails two essential factors: (i) the contact forces at the \ac{cop} act within the friction cone \cite{ellis}, i.e., there is no slipping during the rotation; (ii) there is a torque that enables the rotation around the \ac{cop} in the desired direction.

Inspired by passive walking, we firstly study the physical interaction problem with fully actuated aerial vehicles in a 2-D plane. We now consider a simplified planar system for physical interaction using a symmetric \ac{ee} with two contact points equally distributed around its symmetric axis, see Fig.~\ref{fig:planr}. The whole system is considered as a rigid body. In Fig.~\ref{fig:planr}, with the rigid body assumption, a virtual leg that connects the \ac{cop} and the \ac{com} of the system is equivalent to the leg 1 in Fig.~\ref{fig:passive_walking}. The undesired contact between the \ac{ee} tip and the work surface can thus be seen as \textit{walking} with one leg in contact with the surface. With a desired passive aligning behaviour, the planar system is expected to rotate around the tip of the current contacting leg like a simple pendulum until the other contact point at the \ac{ee} tip touches the surface. For the general 3-D case, in order to achieve the desired passive aligning physical interaction with a rigid and flat work surface using aerial robots, we present three fundamental assumptions of this work:
\begin{itemize}
    \item (1) We denote $\bm{f}_C^n$ as the total normal force vector from the work surface to the system. $\bm{f}_C^t$ presents the total friction force vector parallel to the work surface acting on the system. They both act on the \ac{cop} at contact and the friction cone is applied by:
    \begin{equation}
    |\bm{f}_C^t| \leq \mu_S \cdot |\bm{f}_C^n|,\label{eq:friction_cone}
\end{equation} where $\mu_S \in \mathbb{R}^+$ is the static friction coefficient.
\item (2) A consistent force vector $\bm{f}^B$ (in addition to gravity compensation) along the symmetric axis of the \ac{ee} acts on the \ac{com} of the system towards the work surface.
\item (3) The angular motion in the direction of rotation towards full contact is not actively controlled.
\end{itemize}
The force vector $\bm{f}^B$ in assumption (2) acts as a driving force to generate a torque around \ac{cop} leading to desired full contact. The assumptions (2) and (3) motivate the interaction control design of the studied fully-actuated aerial vehicle in the following sections. The control design is then enhanced by studying the closed loop dynamics in the general 3-D space to ensure this desired behaviour of the platform even in the presence of uncertainties.

\section{modeling and control}\label{sec:control}
\begin{figure*}[t]
      \centering
      \includegraphics[width=0.8\textwidth]{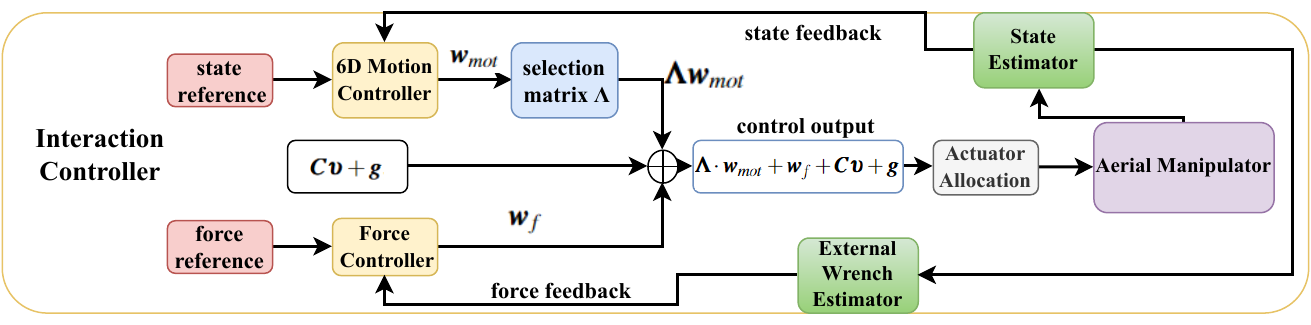}
      \caption{Interaction Control Design }
      \label{fig:controller}
\end{figure*}

In this section, we present the fully-actuated aerial vehicle used in this work and the interaction control design for achieving passive aligning physical interaction based on Sec.~\ref{sec:contact}. We detail the proposed hybrid motion/force controller visualized in Fig.~\ref{fig:controller}. The core idea is to selectively deactivate angular motion control and enable force control in specific directions to generate the driving force $\bm{f}^B$ and make the platform rotate around non-actively controlled axes.
\subsection{System Modeling}
\label{sec:system}
\begin{figure}[t]
   \centering
   \begin{subfigure}{0.48\columnwidth}
\includegraphics[width=\columnwidth]{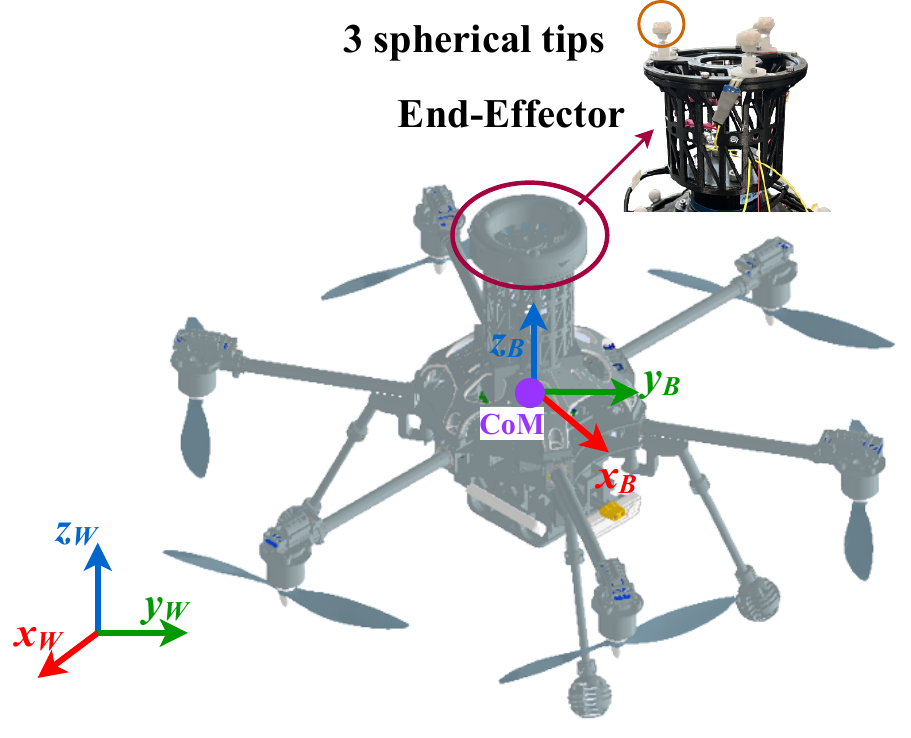}
    \caption{The aerial manipulator}
    \label{fig:frames}
    \end{subfigure} \hfill
    \begin{subfigure}{0.48\columnwidth}
    \includegraphics[width=0.8\columnwidth]{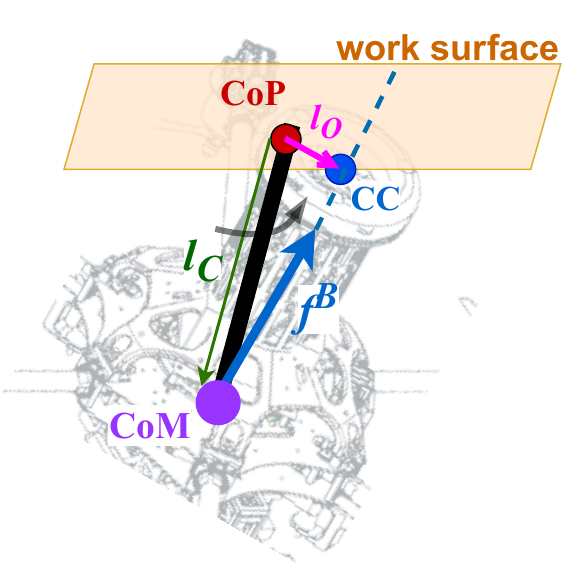}
    \caption{A schematic of the interaction.}
    \label{fig:pendulum}
\end{subfigure}
\caption{(a): Coordinate frames, a fully-actuated aerial vehicle with a cylindrical \ac{ee}. The \ac{ee} has 3 spherical tips on its top surface. (b) Desired passive aligning interaction with the aerial manipulator.}\label{fig:models}
\end{figure}
\begin{figure}[t]
    \centering
\includegraphics[width=0.8\columnwidth]{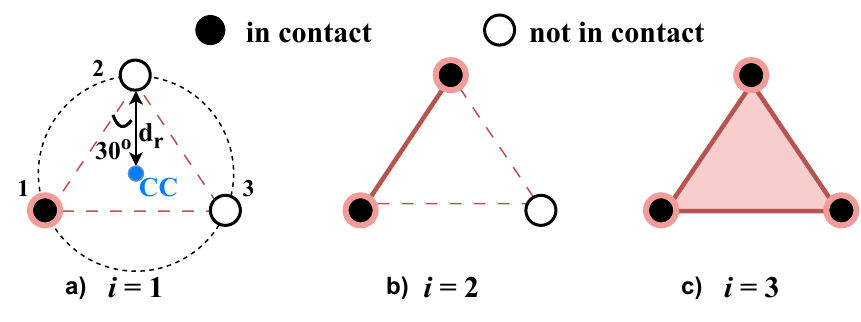}
    \caption{Pink area: support area of $i$ contact points, CC: center of geometry of the \ac{ee} top surface, $d_r$: the radius of the circle where spherical tips locate at with \ac{cc} as the center. The support area gets smaller when $i$ reduces.}
    \label{fig:support_area}
\end{figure}

The aerial manipulator used in this work consists of a fully-actuated aerial vehicle with a rigidly attached cylindrical \ac{ee} as in Fig.~\ref{fig:frames}. The aerial vehicle can generate 6-\ac{dof} wrenches. $\mathcal{F}_W=\{\bm{O}_W;\bm{x}_W,\bm{y}_W,\bm{z}_W\}$ and $\mathcal{F}_B=\{\bm{O}_B;\bm{x}_B,\bm{y}_B,\bm{z}_B\}$ denote the fixed world frame and the body frame rigidly attached to the aerial manipulator \ac{com}, respectively. The axes of $\mathcal{F}_B$ correspond with the principal axes of inertia. The \ac{ee} has three spherical tips mounted on its top surface and is symmetrically attached to the aerial vehicle \ac{wrt} the body axis $\bm{z}_B$ as in Fig.~\ref{fig:frames}. The spherical tips, here called the \textit{feet}, are
spaced 120 degrees apart on a circle of radius $d_r$ \ac{wrt} the \ac{cc} of the \ac{ee} top surface (neglecting the height of the tips), see Fig.~\ref{fig:support_area}. Each \textit{foot} has one single contact point with the work surface due to its spherical shape.
Therefore, the \ac{ee} has $i \in [1,3]$ contact points with the work surface. When $i=3$, the \ac{ee} tip is in full contact with the work surface. The undesired-contact scenario is thus when $i=1,2$. Each of the feet contains a pressure sensor and a spring, with the pressure sensor indicating whether the foot is in contact or not.
Modeling the system as a rigid body, its dynamics can be derived following the Lagrangian formalism expressed in the body frame $\mathcal{F}_B$ as \cite{Bodie2020}:
\begin{equation}\label{eq:em}
\bm{M}\dot{\bm{\upsilon}}+\bm{C}\bm{\upsilon}+\bm{g}=\bm{w}_a+\bm{w}_e,
\end{equation}
where $\bm{\upsilon}=[\bm{\upsilon}_{lin}^\top \ \bm{\upsilon}_{ang}^\top]^{\top} \in \mathbb{R}^{6}$ is the stacked linear and angular velocity, $\bm{M}=diag\big([\bm{M}_{lin} \ \bm{M}_{ang}] \big)$ and $\bm{C} \in \mathbb{R}^{6 \times 6}$ are the  mass and Coriolis matrices, $\bm{g} \in \mathbb{R}^{6}$ is the wrench produced by gravity, $\bm{w}_a$ and $\bm{w}_e \in \mathbb{R}^{6}$ are the actuation wrench and external wrench respectively. Fig.~\ref{fig:pendulum} displays a schematic of the desired passive aligning interaction with the aerial system, where $\bm{f}^B$ is the driving force along $\bm{z}_B$. $\bm{l}_C$ is a vector that points from the \ac{cop} to the system's \ac{com}, and $\bm{l}_0$ points from \ac{cop} to the \ac{cc} of the \ac{ee} tip.
%
%
\subsection{Motion Control} \label{sec:motion_control}
The used fully-actuated aerial vehicle allows 6-\ac{dof} full motion control. The control wrench $\bm{w}_{mot} \in \mathbb{R}^{6}$ of the 6-\ac{dof} motion controller can be derived as:
\begin{equation}\label{eq:motion_control}
\bm{w}_{mot}=\bm{M}\dot{\bm{\upsilon}}_{ref}-\bm{D}_v\bm{e}_v-\bm{K}_p\bm{e}_p,
\end{equation}
where $\dot{\bm{\upsilon}}_{ref} \in \mathbb{R}^{6}$ is the stacked reference linear and angular acceleration, $\bm{D}_v$ and $\bm{K}_p \in \mathbb{R}^{6 \times 6}$ are positive definite matrices representing the damping and stiffness of the system respectively, $\bm{e}_v$ and $\bm{e}_p \in \mathbb{R}^{6}$ are the stacked linear and angular state errors \ac{wrt} the state reference, as in~\cite{Bodie2020}.
\subsection{Force Control}
The force along $\bm{z}_B$ of the body frame is regulated by a proportional-integral (PI) controller using the momentum-based external wrench estimator from~\cite{fabio2014}. We define the force error as $e_f=f_{est}^{z_B}-f_{ref} \in \mathbb{R}$, where $f_{est}^{z_B} \in \mathbb{R}$ is the estimated interaction force component along $\bm{z}_B$ acting on the work surface, and $f_{ref}$ is the force reference. The control output of the direct force controller $\bm{w}_f \in \mathbb{R}^{6}$ can be written as:
\begin{equation}\label{eq:force_control}
    \bm{w}_{f}=\begin{bmatrix}
        \bm{f}^B\\ \bm{0}_3
    \end{bmatrix}=\begin{bmatrix}
        0\\0\\f_{ref}-k_pe_f-k_i\int e_f\\ \bm{0}_3
    \end{bmatrix},
\end{equation}
where $k_p$ and $k_i \in \mathbb{R}^+$ are positive gains.

\subsection{Hybrid motion/force Control}
According to assumptions (2) and (3) in Sec.~\ref{sec:contact}, the proposed hybrid motion/force controller used for interaction involves motion and force control in selected axes by combining the control wrenches in Eq.(\ref{eq:motion_control})(\ref{eq:force_control}), as in Fig.~\ref{fig:controller}. The final feedback-linearized actuation wrench $\bm{w}_a \in \mathbb{R}^{6}$ is defined as:
\begin{equation}
    \bm{w}_a=\bm{\Lambda}\cdot \bm{w}_{mot}+\bm{w}_{f}+\bm{C}\bm{\upsilon}+\bm{g}, \label{eq:control_output}
\end{equation}
where
$\bm{\Lambda}=diag(\begin{bmatrix}1 &1& 0& 0& 0& 1\end{bmatrix})$ is the selection matrix which eliminates the linear motion control in $\bm{z}_B$  and the angular motion control in $\bm{x}_B$ and $\bm{y}_B$ from $\bm{w}_{mot}$. With the actuation wrench in Eq.~(\ref{eq:control_output}), the closed loop system dynamics yields to:
\begin{equation}
    \bm{M}\dot{\bm{\upsilon}}=\bm{\Lambda}\bm{w}_{mot}+\bm{w}_{f}+\bm{w}_{e}. \label{eq:close_loop}
\end{equation}
With this control design, the system dynamics passively responds to the external wrenches in the selected non-actively controlled directions.

\section{Passive Dynamics of the System}
\label{sec:passive_dynamics}
The 6-\ac{dof} motion controller in Sec.\ref{sec:motion_control} is used during free flight to reach a reference position close to the work surface. After reaching the reference position, we manually enable the interaction controller in Fig.~\ref{fig:controller} with the initial force reference $f_{ref}$ being zero for smooth transition between controllers. For approaching and aligning with the work surface, we command the system with a reference force $f_{ref}>0$. We investigate the passive aligning behaviour of the studied platform in detail in Sec.~\ref{sec:contact}. 

\subsection{Numerical Indicators of Contact Status}
\begin{figure}[t]
    \centering  
    \begin{subfigure}{0.24\textwidth}  
    \includegraphics[width=0.8\columnwidth]{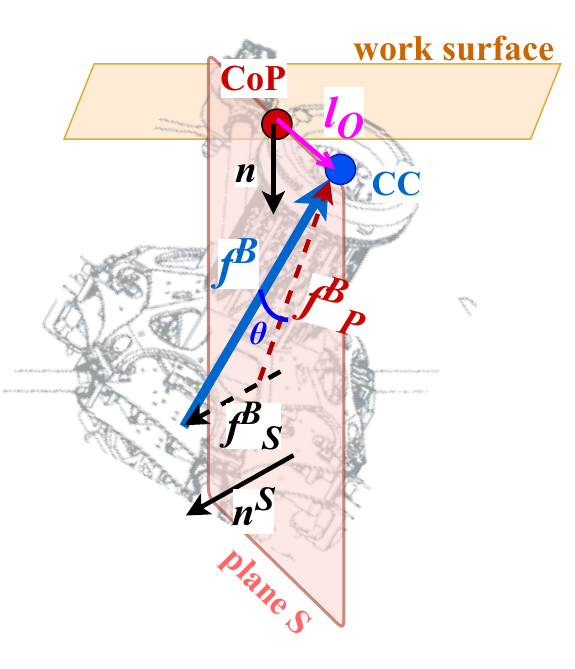}
    \caption{}\label{fig:f_B}
    \end{subfigure}
     \begin{subfigure}{0.23\textwidth}  
    \includegraphics[width=0.8\columnwidth]{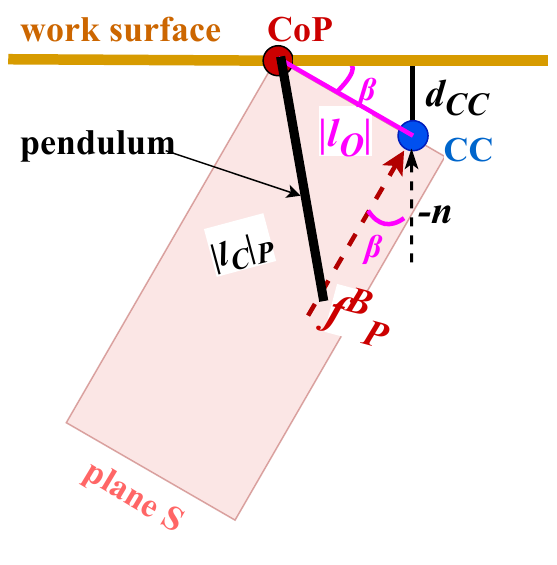}
    \caption{ }\label{fig:splane}
    \end{subfigure}
    \caption{a) $\bm{f}^B=\bm{f}^B_{\mathbb{S}}+\bm{f}^B_P$, decomposed \ac{wrt} the 2-D plane $\mathbb{S}$ constructed by vectors $\bm{n}$ and $\bm{l}_O$, $\bm{n}^{\mathbb{S}}$ is the normal unit vector of plane $\mathbb{S}$, and $\theta$ denotes the angle between $\bm{f}^B$ and $\bm{f}^B_P$; b) front view of $\mathbb{S}$ plane with projected $|\bm{l}_C|$: $|\bm{l}_C|_P$, $\beta$ is the angle from $\bm{-n} $ to $\bm{f}^B_P$.}
    \label{fig:decompose}
\end{figure}

To achieve the desired passive aligning interaction in the presence of uncertainties, we propose quantitative indicators to evaluate the contact status between the \ac{ee} tip and the work surface. With these indicators, we derive two essential conditions to be met to ensure the desired passive aligning interaction. 
We define $d_{CC} \in \mathbb{R}^+$ as the distance from the \ac{cc} of the \ac{ee} top surface to the work surface along the normal unit vector $\bm{n}$ of the work surface — a measure of contact status. When $d_{CC}=0$, the \ac{ee} tip is fully in contact with the work surface. $d_{CC}\neq0$ signifies undesired contact. We aim at bringing $d_{CC}$ to zero with the torque generated around the \ac{cop} by $\bm{f}^B$. To analyze the effect of $\bm{f}^B$ on $d_{CC}$, we decompose $\bm{f}^B$ into two components as in Fig.~\ref{fig:f_B} denoted by:
\begin{equation}\label{eq:decomp}
    \bm{f}^B=\bm{f}^B_{\mathbb{S}}+\bm{f}^B_P,
\end{equation}
 where $\mathbb{S}$ is a 2-D plane constructed by $\bm{n}$ and $\bm{l}_O$, $\bm{f}^B_P$ is the projected $\bm{f}^B$ on the plane $\mathbb{S}$, and $\bm{f}^B_{\mathbb{S}}$ denotes the component along the normal unit vector $\bm{n}^{\mathbb{S}}$ of the plane $\mathbb{S}$. We define $\theta \in [0, 90\degree)$ as the angle between $\bm{f}^B$ and $\bm{f}^B_P$, and $\beta \in (-90\degree,90\degree)$ as the angle from $-\bm{n}$ to $\bm{f}^B_P$.
Based on the definition of $d_{CC}$ and the front view of the plane $\mathbb{S}$ in Fig.~\ref{fig:splane}, the relationship between $d_{CC}$ and $\beta$ is given by:
\begin{equation}\label{eq:dcc}
    d_{CC}=|\bm{l}_O|sin(|\beta|),
\end{equation}
with which, the angle $\beta$ is directly related to the contact status. Considering the support area in Fig.~\ref{fig:support_area}, $|\bm{l}_O|$ can vary among:
\begin{equation}
 \begin{cases}
   |\bm{l}_O|=d_r & \text{for $i=1$,}\\
     \frac{d_r}{2} \leq|\bm{l}_O|\leq d_r, & \text{for $i=2$},\\
    0 \leq|\bm{l}_O|\leq d_r, &\text{for $i=3$}.
  \end{cases}\label{eq:lo}
\end{equation}
Therefore, if and only if $d_{CC}=\beta=0$ where $i=3$, $|\bm{l}_O|$ can be zero. The similarity between $d_{CC}$ and $\beta$ makes $\beta$ a good proxy for representing the contact status with $\beta=0$ being the full-contact scenario. Furthermore, we study the dynamics within the $\mathbb{S}$ plane, which is key for understanding how $\bm{f}^B$ affects contact status across the entire system. In the following sections, we outline the conditions to meet the fundamental assumptions in Sec.~\ref{sec:conclusion} for achieving passive aligning physical interactions, even in the presence of uncertainties.
\begin{figure}[t]
    \centering   \includegraphics[width=0.8\columnwidth]{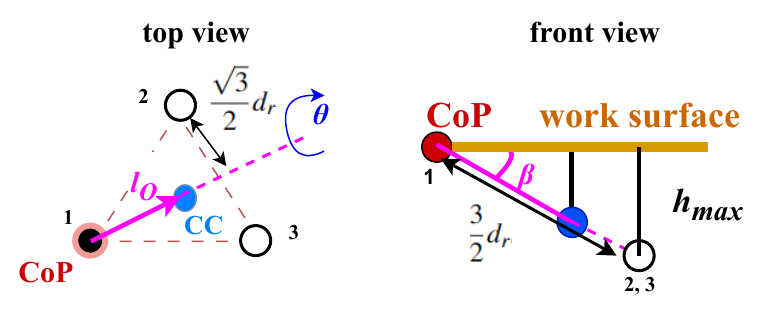}
    \caption{When $\theta=0$, $h_{max}=\frac{3}{2}d_r sin(|\beta|)$ is the maximum allowed height change of foot 2 or 3 before touching the work surface with increased $\theta$.}
    \label{fig:theta}
\end{figure}
\subsection{Friction-Ensuring Condition}
In this section, we derive the mathematical expression which ensures the friction cone assumption in Sec.~\ref{sec:contact}. When the platform is in the undesired-contact scenario, with the decomposition of $\bm{f}^B$ in Eq.(\ref{eq:decomp}), the resulted contact forces $|\bm{f}_C^n|$ and $|\bm{f}_C^t|$ introduced in Sec.\ref{sec:contact} can be written as:
\begin{equation}\label{eq:mu}
    \begin{split}
&|\bm{f}_C^n|=|\bm{f}^B_P|cos(\beta),\\
    &|\bm{f}_C^t|=\sqrt{(|\bm{f}^B_P|sin(|\beta|))^2+(|f^B_{\mathbb{S}}|)^2},\\
    & \mu= \frac{|\bm{f}_C^t|}{|\bm{f}_C^n|},
    \end{split}
\end{equation}
where as in Fig.~\ref{fig:f_B},
\begin{equation*}
    |\bm{f}^B_P|=|\bm{f}^B|cos(\theta), \ |f^B_{\mathbb{S}}|=|\bm{f}^B|sin(\theta).
\end{equation*}
The above equations accurately describe the forces in the general 3-D space but add complexity in the mathematical analysis later. Therefore, to simplify the problem, we assume that when $i=2$, the \ac{cop} locates at the center of the line in Fig.~\ref{fig:support_area} (b) which is the ideal case when only $\bm{f}^B$ is acting on the \ac{com} in addition to gravity compensation. With this assumption, $\bm{f}^B$ lies in the plane $\mathbb{S}$, which yields the following simplification:
\begin{align}
    |\bm{l}_O|=\frac{d_r}{2}, & \ \theta=0, &\text{for $i=2$.}
\end{align}
During $i=1$, an increased $\theta$ value will rotate the \ac{ee} top surface around $\bm{l}_O$ with a radius of $\frac{\sqrt{3}}{2}d_r$ until the second \textit{foot} touches the work surface, see Fig.~\ref{fig:theta}. The geometric relations for $\theta=0$ are shown in Fig.~\ref{fig:theta}. The height change of foot 2 or 3 caused by the increased $\theta$ is limited by the work surface which introduces an upper bound to the maximum value of $\theta$, here we call $\theta_{max}$. At $\theta=0$, the maximum height change $h_{max}$ of foot 2 or 3 and the upper bound of $\theta_{max}$ can be derived as:
\begin{align}
    &h_{max}=\frac{3}{2}d_r sin(|\beta|), \
    &sin(\theta_{max})<\frac{h_{max}}{\frac{\sqrt{3}d_r}{2}}=\sqrt{3}sin(|\beta|).\label{eq:max_theta}
\end{align}

With Eq.~\ref{eq:max_theta}, the ratio $\mu$ in Eq.~\ref{eq:mu} equates to:
\begin{equation} \label{eq:mulim}
\mu=\frac{|\bm{f}_C^t|}{|\bm{f}_C^n|} <  tan(|\beta|)\sqrt{1+\frac{3}{1-3sin^2(|\beta|)}}=\mu_{lim}(|\beta|),
\end{equation}
where $\mu_{lim}$ is an upper boundary of $\mu$.
The $\mu_{lim}$ as a function of $|\beta|$ is visualized in Fig.\ref{fig:mu} which shows that $\mu_{lim}$ increases along with $|\beta|$ in the range $[0,30\degree]$. $|\beta|$ varies within a range due to perturbations caused by uncertainties with $|\beta|_{max}$ being the maximum value in the range. Based on Eq.(\ref{eq:mulim}), if the friction condition in Eq.(\ref{eq:friction_cone}) holds for the upper boundary $\mu_{lim}$ with $|\beta|=|\beta|_{max}$, then the friction condition holds for the entire range of $|\beta|$.
 Therefore, to ensure the friction condition in Eq.(\ref{eq:friction_cone}), we propose the following condition:
\begin{itemize}
    \item \textbf{Condition 1} 
    \begin{equation}
    \mu_{lim}(|\beta|_{max}) \leq \eta\mu_S,\label{eq:necessary}
\end{equation}
\end{itemize}
with $\eta \in (0,1]$ being a safety factor used to account for for uncertainties that may affect the contact forces.
\subsection{Rotation-Ensuring Condition}
 When Eq.(\ref{eq:friction_cone}) holds, we derive the condition to be met that guarantees the desired rotation towards full contact by studying the passive dynamics in the plane $\mathbb{S}$. We define $\tau_d \in \mathbb{R}$ as the perturbation of rotational dynamics caused by uncertainties (neglecting the short impact period during the initial contact). We denote $|\bm{l}_C|_P$ as the projected length of $|\bm{l}_C|$ in plane $\mathbb{S}$. The simplified pendulum dynamics in plane $\mathbb{S}$ can be written as:
\begin{equation}
    \label{eq:pendulum}
    m|\bm{l}_C|_P^2\ddot{\beta}=\tau_e^p-\tau_d,
\end{equation}
where
\begin{equation}\label{eq:tau_ep}
|\tau_e^p|=|\bm{f}^B_P||\bm{l}_O|,
\end{equation}
with $m$ being the mass of the whole system, and $\tau_e^p \in \mathbb{R}$ is the torque generated by $\bm{f}^B_P$ \ac{wrt} the reference point \ac{cop}. With Eq.(\ref{eq:tau_ep}) the system rotates with an angular acceleration of:
\begin{equation}
\Ddot{\beta}=\frac{1}{m|\bm{l}_C|_P^2}(|\bm{f}^B_P||\bm{l}_O|-\tau_d).
    \end{equation}
Knowing that $|\bm{f}^B_P|=|\bm{f}^B|cos(\theta)\geq |\bm{f}^B|cos(\theta_{max})$ and $|\bm{l}_O|\geq \frac{d_r}{2}$ for $i=1,2$ and considering Eq.(\ref{eq:max_theta}), one has:
\begin{equation}
    |\tau_e^p|\geq |\bm{f}^B| \cdot cos(\theta_{max})\frac{d_r}{2} > |\bm{f}^B|\sqrt{1-3sin^2(
|\beta|_{max})}\frac{d_r}{2}.
\end{equation}

To ensure that  $\frac{\ddot{\beta}}{|\ddot{\beta}|}=-\frac{\beta}{|\beta|}$ towards reducing $|\beta|$ under uncertainties, we propose a second condition to be met where:
\begin{itemize}
    \item \textbf{Condition 2}
    \begin{equation}\label{eq:c2}
    |\bm{f}^B|\sqrt{1-3sin^2(
|\beta|_{max})}\frac{d_r}{2} >|\tau_d|_{max},
    \end{equation}
\end{itemize}
and $|\tau_d|_{max}$ is the maximum value of $|\tau_d|$.
\subsection{Design Guidelines}
The presented two conditions can serve as valuable guidelines for hardware as well as control design. \textbf{Condition 1} provides the minimum $\mu_S$ required to ensure the friction cone assumption in Sec.~\ref{sec:contact} against uncertainties which simplifies the selection of contact materials at the \ac{ee} tip. \textbf{Condition 2} can be used to design the \ac{ee} size knowing the preferred force value $|\bm{f}^B|$ to be generated from the platform, or to calculate the minimum force required knowing the size of the EE, to ensure the desired rotation in the presence of uncertainties. In the following section, we present practical experiments that validate the control approach and the guidelines.

\section{Experiments}\label{sec:experiments}
In this section, we show four groups of experiments with different ranges of $|\beta|$ and friction coefficients $\mu_S$ between the \ac{ee} tip and the work surface to validate the functionality of \textbf{Condition 1}. Moreover, experiments with different force reference $f_{ref}$ are executed with the same $|\beta|$ range and $\mu_S$ to validate the effectiveness of \textbf{Condition 2}.
\subsection{Experiments Setup}
\begin{figure}[t]
    \centering   
    \begin{subfigure}{0.23\textwidth}
    \includegraphics[width=\columnwidth]{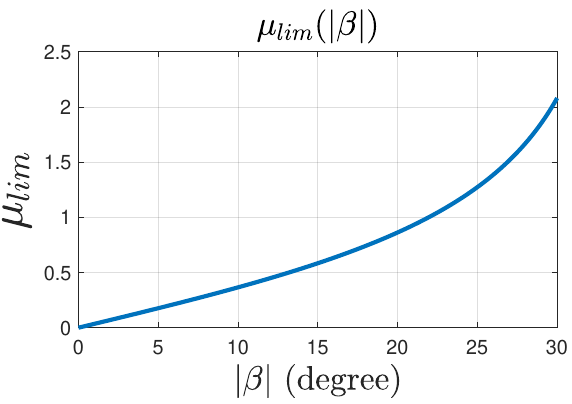}
    \caption{}
    \label{fig:mu}
    \end{subfigure}
    \hfill
    \begin{subfigure}{0.24\textwidth}
    \includegraphics[width=\columnwidth]{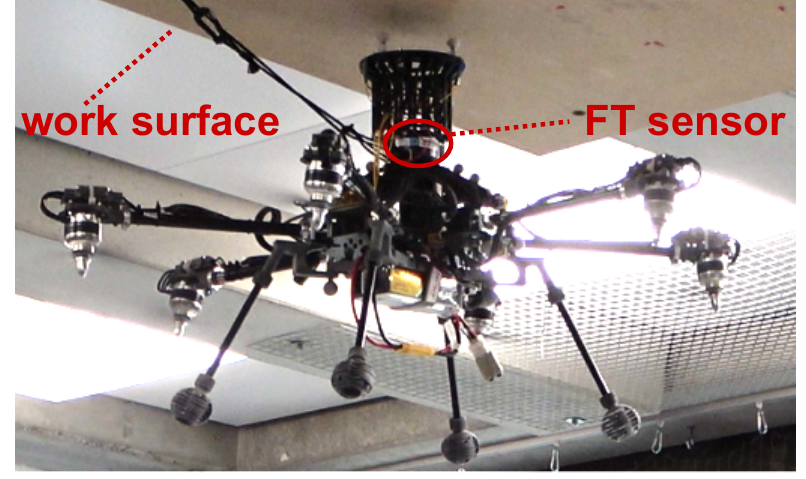}
    \caption{}
    \label{fig:setup}
    \end{subfigure}
    \caption{(a): $\mu_{lim}$ increases along with $|\beta|$. (b): Aerial robot in interaction with the work surface mounted on the ceiling.}
\end{figure}

The aerial manipulator described in Sec.~\ref{sec:system} is used for all experiments. A flat wooden board mounted on the ceiling is set up as the work surface, see Fig.~\ref{fig:setup}. A Vicon motion capture system is used for obtaining position and orientation. Additionally, a 6-\ac{dof} FT (force and torque) sensor is mounted below the \ac{ee} to measure the external forces and torques during interaction. The purpose of using the FT sensor is to obtain ground truth contact information for the system's dynamics study, and is not used for force control. Pressure sensors at the \ac{ee} tip are used to check the contact status of the three \textit{feet}. Considering the dimensions of the platform, the \ac{ee} is designed with $d_r=\SI{0.0525}{\meter}$. 
\subsection{Verification of \textbf{Condition 1}}
\begin{figure}[t]
   \centering
   \begin{subfigure}{0.24\textwidth}
\includegraphics[width=\textwidth]{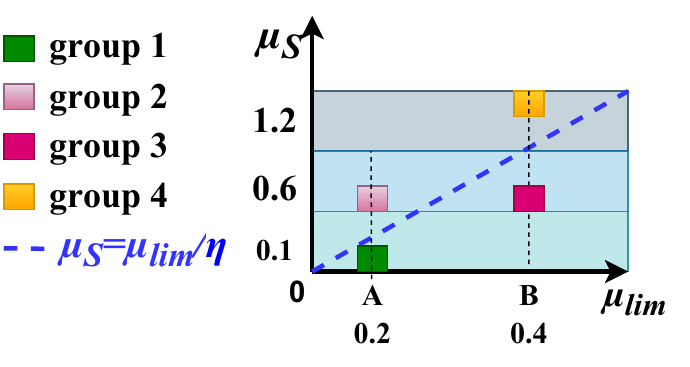}
    \caption{}
    \label{fig:friction}
\end{subfigure}
\hfill
\begin{subfigure}{0.23\textwidth}
\includegraphics[width=\textwidth]{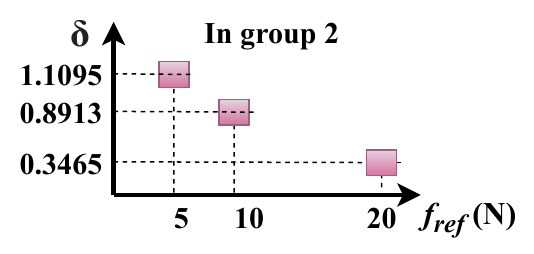}
    \caption{}
    \label{fig:rsme}
    \end{subfigure}\caption{a) four groups of experiments with different ranges of $|\beta|$ as (A) and (B), under different friction coefficients $\mu_S$, the dashed line is the minimum $\mu_S$ to meet \textbf{Condition 1}; b) experiments in group 2 with range (A) and $\mu_S \approx 0.6$ subjected to different force reference $f_{ref}$ sent to the force control, $\delta$ is the evaluator of contact status using pressure sensor data, lower $\delta$ presents better contact.}
\end{figure}
\begin{figure}[t]
    \centering   \includegraphics[width=\columnwidth]{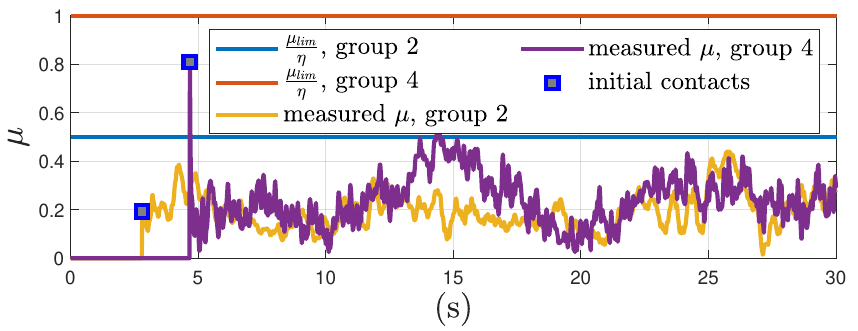}
    \caption{Measured $\mu$ from the FT sensor measurements, $\frac{\mu_{lim}}{\eta}$ is the minimum $\mu_S$ to meet \textbf{Condition 1}. The measured $\mu$ of group 2 and 4 is always smaller than $\frac{\mu_{lim}}{\eta}$ indicating satisfied \textbf{Condition 1}.}
    \label{fig:mu_meas}
\end{figure}

The four groups of experiments presented in this section are shown in Fig.~\ref{fig:friction} which involve two ranges of $|\beta|$ value and three ranges of static friction coefficients at contact. Considering the above experimental setup, we arranged experiments in two ranges of $|\beta|$ value as: range (A) - $[0,6\degree]$, and range (B) - $[7,11\degree]$. From these two ranges of $|\beta|$, we can calculate the corresponding $\mu_{lim}$, as $\mu_{lim}^A=\mu_{lim}(|\beta|=6\degree)=0.2$, and $\mu_{lim}^B=\mu_{lim}(|\beta|=11\degree)=0.4$. To ensure \textbf{Condition 1} under uncertainties, we choose $\eta=0.4$. Therefore the static friction coefficients that can ensure friction cone assumption for range (A) and (B) are: $\mu_S \geq \frac{\mu_{lim}^A}{\eta}=0.5$, and $\mu_S \geq \frac{\mu_{lim}^B}{\eta}=1$. Three different approximate friction coefficients between the \ac{ee} tip and the work surface are tested, where case 1 has $\mu_S \approx 0.1$, case 2 has $\mu_S \approx 0.6$, and case 3 has $\mu_S \approx 1.2$. The static friction coefficients are regulated by changing the materials at the \ac{ee} tip and the work surface. Group 1 and group 2 are tested under range (A) but with friction conditions of case 1 and 2 respectively. Group 3 and group 4 are tested under range (B) but with friction conditions of case 2 and 3 respectively. The dashed line in Fig.~\ref{fig:friction} presents the case when $\mu_S$ equals to the minimum value $\frac{\mu_{lim}}{\eta}$ to meet \textbf{Condition 1}. Condition 1 is intentionally not satisfied in group 1 and 3 and we therefore expect the system to slip and become unstable, while it is met for group 2 and 4. The failure cases of group 1 and 3 are shown in the attached video in which the platform slipped and instability occurred. The platform accomplished the interaction task in group 2 and 4. The measured force ratio $\mu=\frac{|\bm{f}_C^t|}{|\bm{f}_C^n||}$ from the FT sensor for group 2 and 4 is shown in Fig.~\ref{fig:mu_meas}. From the failure of group 1 and 3, we can conclude that ensuring the friction cone assumption is crucial for stable interaction. Moreover, the force ratio measurements from group 2 and 4 confirm the expected friction condition during the interaction, see Fig.~\ref{fig:mu_meas}.
\begin{figure}[t]
    \centering   \includegraphics[width=\columnwidth]{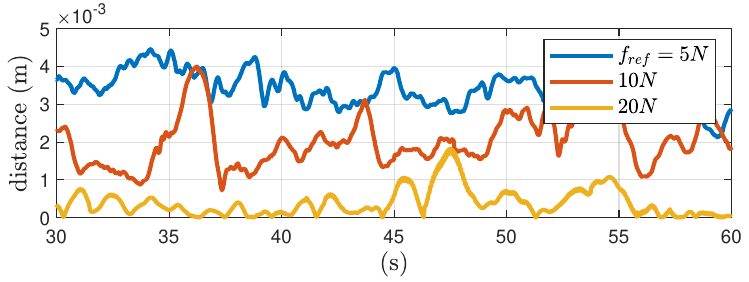}
    \caption{Tests in group 2: $d_{CC}$ for different $f_{ref}$. When $f_{ref}$ increases, $d_{CC}$ reduces indicating better contact status. At $f_{ref}=20$\si{\newton}, $d_{CC}$ has the lowest value.}
    \label{fig:dcc}
\end{figure}
\subsection{Verification of \textbf{Condition 2}}
In this section, we validate the effectiveness of \textbf{Condition 2} through three tests in group 2 with different force reference $f_{ref}$ sent to the force control. In group 2, range (A) of $|\beta| \in [0,6\degree]$ and static friction coefficient of $\mu_S \approx 0.6$ are used, as in Fig.~\ref{fig:friction}. The rotational torque caused by uncertainties $|\tau_d|$ can be estimated via the external wrench estimation while hovering during free flight, which varies in a range of $[0, \ 0.5]$\SI{}{\newton\meter} according to the test results. With the setup in group 2, \textbf{Condition 2} yields: $|\bm{f}^B|\sqrt{1-3sin^2(
6\degree)}\frac{0.0525}{2}>0.5$ which results into $|\bm{f}^B|>19.4$\si{\newton}. Therefore, to ensure the desired rotational dynamics with the current setup, it is suggested to set the reference force magnitude $f_{ref}$ to at least $20$\si{\newton}.
To quantitatively evaluate the contact status of the three feet with different $f_{ref}$ values as $f_{ref}=$\SI{5}{\newton}, \SI{10}{\newton}, \SI{20}{\newton}, a numerical evaluator $\delta$ is defined as:
\begin{equation}
    \delta=\frac{1}{3}\sum _{j=1}^{3}RSME(\frac{\bm{S}^j_{1\times n}}{S_{ref}},\bm{I}_{1\times n}),
\end{equation}
where $\bm{S}^j_{1\times n}$ denotes the sequence of the $j-th, j=1,2,3$ pressure sensor measurements during physical interaction, and 
$\bm{I}_{1\times n}$ is an array of ones with the same size. $S_{ref}=\frac{1}{3}|f^{Z_B}_{meas}|$ is the desired value of each pressure sensor, where $f^{Z_B}_{meas}$ is the measured force value along $\bm{z}_B$ by the FT sensor during the interaction. $RMSE(\bm{a},\bm{b})$ represents the Root-Mean-Square-Error between the arrays $\bm{a}$ and $\bm{b}$. $\delta$ thus denotes the average RMSE of three pressure sensors. The $\delta$ value from the tests in group 2 is shown in Fig.~\ref{fig:rsme}. Ideally the three pressure sensors should equally contribute to the total force value with $\delta=0$. Therefore, a smaller value of $\delta$ indicates a better contact status between the \ac{ee} tip and the work surface. Among group 2, with $f_{ref}=20$\si{\newton}, we obtained the best contact status with the lowest $\delta$ value as in Fig.~\ref{fig:rsme}. The measurements of $d_{CC}$ in Fig.~\ref{fig:dcc} indicate that $f_{ref}=20$\si{\newton} establishes a better contact at steady-state with smaller magnitude of $d_{CC}$. With the above experiments, we validated the effectiveness of \textbf{Condition 2}. 

Additionally, we present a set of interaction experiments with work surfaces at various orientation, as in Fig.~\ref{fig:action}. Executed experiments are displayed in the attached video, also available at \url{https://youtu.be/AMHJiGw5r6E}.

\section{Conclusion and Discussion}\label{sec:conclusion}
In this paper, we introduced a hybrid motion/force control strategy augmented by passive dynamics. We provided a detailed analysis of the critical conditions that guarantee passive aligning physical interactions with fully-actuated aerial vehicles. These contributions serve as practical design guidelines for both hardware and control of future aerial manipulators, and address the practical challenges associated with the utilization of fully-actuated aerial vehicles in NDT applications. We verified the proposed control strategy and design guidelines through extensive experiments. The proposed guidelines offer a solution that allows us to achieve the targeted interaction task without adding excessive complexity to the control design. In the existing literature of aerial manipulation, forces and torques coming from the environment acting on the robotic system are often compensated by active control as disturbances. In this work, we showed that in some cases, the external wrench generated by contact forces can be used to assist the system in accomplishing interaction tasks. A comprehensive understanding of the contact forces and of the contact conditions is key to identify these cases.

\addtolength{\textheight}{-12cm}   






% ...

\end{document}